\documentclass[10pt,twocolumn,letterpaper]{article}

\usepackage{cvpr}

\usepackage{epsfig}
\usepackage{graphicx}
\usepackage{amsmath}
\usepackage{amssymb}
\usepackage{bm}
\usepackage{multirow}
\usepackage{multicol}
\usepackage{colortbl,hhline}
\usepackage{booktabs}
\usepackage{array,eucal}

\usepackage{authblk}

\usepackage[margin=4pt,font=small,labelfont=bf,labelsep=endash,tableposition=top]{caption}

\usepackage[pagebackref=true,breaklinks=true,letterpaper=true,colorlinks,bookmarks=false]{hyperref}

\cvprfinalcopy %

\begin{document}

\title{DoDNet: Learning to segment multi-organ and tumors \\
from multiple partially labeled datasets
}

\author[1,2]{\rm Jianpeng Zhang\thanks{JZ and YX contributed equally. Work was done when JZ and YX were visiting The University of Adelaide.}}
\author[1,2]{\rm Yutong Xie$^*$}
\author[1]{\rm Yong Xia}
\author[2]{\rm Chunhua Shen}
\affil[1]{~School of Computer Science and Engineering, Northwestern Polytechnical University, China}
\affil[2]{~The University of Adelaide, Australia}
\affil[ ]{{\tt\small \{james.zhang, xuyongxie\}@mail.nwpu.edu.cn}; {\tt\small yxia@nwpu.edu.cn}; {\tt\small chunhua.shen@adelaide.edu.au}}

\maketitle
\begin{abstract}

Due to the intensive cost of labor and expertise in annotating 3D medical images at a voxel level, most benchmark  datasets are equipped with the annotations of only one type of organs and/or tumors, resulting in the so-called partially labeling issue. To address this, 
we propose a dynamic on-demand network (DoDNet) that learns to segment multiple organs and tumors on partially labeled datasets. 

DoDNet consists of a shared encoder-decoder architecture, a task encoding module, a controller for %
generating dynamic convolution filters,
and a single but dynamic segmentation head. 
The information of the current segmentation task is encoded as a task-aware prior to tell the model what the task is expected to %
solve. 
Different from existing approaches which fix kernels after training, the kernels in dynamic head are generated adaptively by the controller, conditioned on both input image and assigned task. Thus, DoDNet is able to segment multiple organs and tumors, as done by multiple networks or a multi-head network, in a much efficient and flexible manner. 
We have created a large-scale partially labeled dataset, termed MOTS, and demonstrated the superior performance of our DoDNet over other competitors on seven organ and tumor segmentation tasks. We also transferred the weights pre-trained on MOTS to a downstream multi-organ segmentation task and achieved state-of-the-art performance.
This study provides a general 3D medical image segmentation model that has been pre-trained on a large-scale partially labelled dataset and can be extended (after fine-tuning) to downstream volumetric medical data segmentation tasks. 
The dataset and code 
are available at:   
  \def\UrlFont{\rm\small\ttfamily}
\url{https://git.io/DoDNet}

\end{abstract}

\section{Introduction}

\begin{figure}[t]
\centering 
\includegraphics[width=0.9\linewidth]{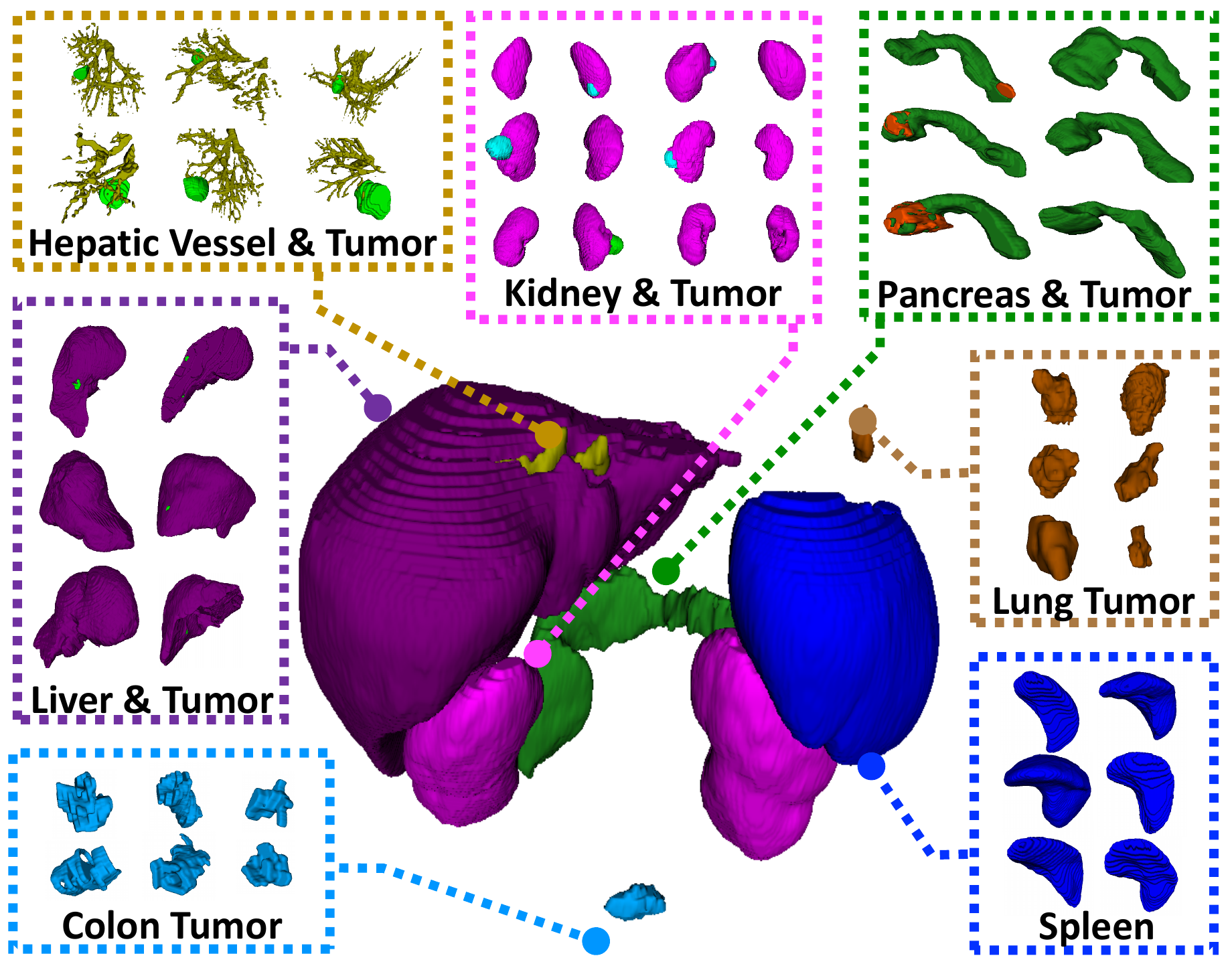}
\caption{
Illustration of partially labeled multi-organ and tumor segmentation. This task aims to segment multiple organs and tumors using a network trained on several partially labeled datasets, each of which is originally specialized for the segmentation of a particular abdominal organ and$/$or related tumors. For instance, the first dataset only has annotations of the liver and liver tumors, and the second dataset only provides annotations of kidneys and kidney tumors. Here each color represents a partially labeled dataset.}
\label{fig:fig1}
\end{figure}

Automated segmentation of abdominal organs and tumors using computed tomography (CT) is one of the most fundamental yet challenging tasks in medical image analysis \cite{pang2020ctumorgan,kavur2020chaos}. It plays a pivotal role in a variety of computer-aided diagnosis tasks, including lesion contouring, surgical planning, and 3D reconstruction.
Constrained by the labor cost and expertise, it is hard to annotate multiple organs and tumors at voxel level in a large dataset. Consequently, most benchmark datasets were collected for the segmentation of only one type of organs and$/$or tumors, and all task-irrelevant organs and tumors were annotated as the background (see Fig.~\ref{fig:fig1}). For instance, the LiTS dataset \cite{bilic2019liver} only has annotations of the liver and liver tumors, and the KiTS dataset \cite{heller2019kits19} only provides annotations of kidneys and kidney tumors. These partially labeled datasets are distinctly different from the segmentation benchmarks in other computer vision areas, such as PASCAL VOC \cite{everingham2010pascal} and Cityscapes \cite{cordts2016cityscapes}, where multiple types of objects were annotated on each image.
Therefore, one of the significant challenges facing multi-organ and tumor segmentation is the so-called {\textit{partially labeling issue}}, \textit{i.e.}, how to learn the representation of multiple organs and tumors under the supervision of these partially annotated images.

\begin{figure*}[t]
\begin{center}
\includegraphics[width=1.0\linewidth]{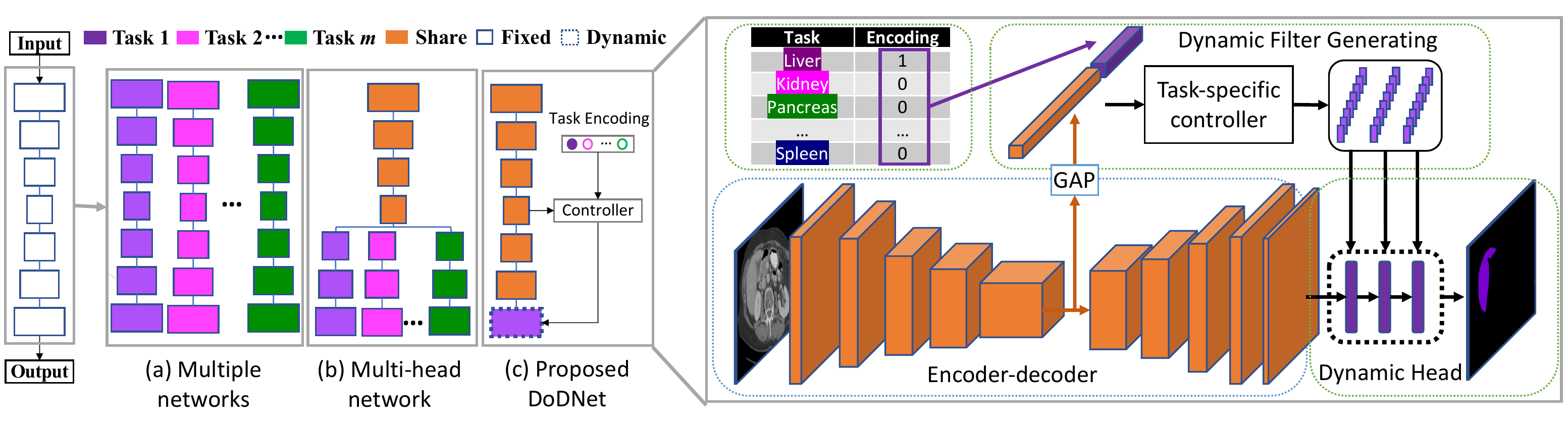}
\end{center}
\caption{Three types of methods to perform $m$ partially labeled segmentation tasks.
(a) Multiple networks: Training $m$ networks on $m$ partially labeled subsets, respectively;
(b) Multi-head networks: Training one network that consists of a shared encoder and $m$ task-specific decoders (heads), each performing a partially labeled segmentation task; and
(c) Proposed DoDNet: It has an encoder, a task encoding module, a dynamic filter generation module, and a dynamic segmentation head. The kernels in the dynamic head are conditioned on the input image and assigned task.}
\label{fig:framework}
\end{figure*}

Mainstream approaches address this issue via separating the partially labeled dataset into several fully labeled subsets and training a network on each subset for a specific segmentation task \cite{yu2019crossbar,isensee2019automated,zhang2019light,myronenko20193d,zhu2019multi}, resulting in `multiple networks' shown in Fig.~\ref{fig:framework}(a). Such an intuitive strategy, however, increases the computational complexity dramatically.
Another commonly-used solution is to design a multi-head network (see Fig.~\ref{fig:framework}(b)), which is composed of a shared encoder and multiple task-specific decoders (heads) \cite{chen2019med3d,fang2020multi,shi2020marginal}. In the training stage, when each partially labeled data is fed to the network, only one head is updated and others are frozen. The inferences made by other heads are unnecessary and wasteful. Besides, the inflexible multi-head architecture is not easy to extend to a newly labeled task.

In this paper, we propose a dynamic on-demand network (DoDNet), which can be trained on partially labeled datasets for multi-organ and tumor segmentation.
DoDNet is an encoder-decoder network with a single but dynamic head (see Fig.~\ref{fig:framework}(c)), which is able to segment multiple organs and tumors as done by multiple networks or a multi-head network.
The kernels in the dynamic head are generated adaptively by a controller, conditioned on the input image and assigned task. Specifically, the task-specific prior is fed to the controller to guide the generation of dynamic head kernels for each segmentation task.
Owing to the light-weight design of the dynamic head, the computational cost of repeated inference can be ignored when compared to that of a multi-head network. 
We evaluate the effectiveness of DoDNet on seven organ and tumor segmentation benchmarks, involving the liver and tumors, kidneys and tumors, hepatic vessels and tumors, pancreas and tumors, colon tumors, and spleen. Besides, we transfer the weights pre-trained on partially labeled datasets to a downstream multi-organ segmentation task, and achieve state-of-the-art performance on the Multi-Atlas Labeling Beyond the Cranial Vault Challenge dataset. 
Our contributions are three-fold:
\begin{itemize}
\item We attempt to address the partially labeling issue from a new perspective, \textit{i.e.}, proposing a single network that has a dynamic segmentation head to segment multiple organs and tumors as done by multiple networks or a multi-head network.
\item Different from the traditional segmentation head which is fixed after training, the dynamic segmentation head in our model is adaptive to the input and assigned task, leading to much improved efficiency and flexibility.
\item The proposed DoDNet pre-trained on partially labeled datasets can be transferred to downstream annotation-limited segmentation tasks, and hence is beneficial for the medical community where only limited annotations are available for 3D image segmentation.%
\end{itemize}

\section{Related Work}
{\noindent\bf Partially labeled medical image segmentation}
Segmentation of multiple organs and tumors is a generally recognized difficulty in medical image analysis \cite{Xie_multiorgan,zhang2020block,wang2019abdominal,schoppe2020deep}, particularly when there is no large-scale fully labeled datasets. Although several partially labeled datasets are available, each of them is specialized for the segmentation of one particular organ and$/$or tumors.
Accordingly, a segmentation model is usually trained on one partially labeled dataset, and hence is only able to segment one particular organ and tumors, such as the liver and liver tumors \cite{li2018h, zhang2019light, seo2019modified, tang2020net}, kidneys and kidney tumors \cite{myronenko20193d, Hou_kidney_isbi}. Training multiple networks, however, suffers from the waste of computational resources and a poor scalability. 

To address this issue, many attempts have been made to explore multiple partially labeled datasets in a more efficient manner. 
Chen \textit{et al.} \cite{chen2019med3d} collected multiple partially labeled datasets from different medical domains, and co-trained a heterogeneous 3D network on them, which is specially designed with a task-shared encoder and task-specific decoders for eight segmentation tasks. 
Huang \textit{et al.} \cite{huang2020multi} proposed to co-train a pair of weight-averaged models for unified multi-organ segmentation on few-organ datasets. 
Zhou \textit{et al.} \cite{zhou2019prior} first approximated anatomical priors of the size of abdominal organs on a fully labeled dataset, and then regularized the organ size distributions on several partially labeled datasets. 
Fang \textit{et al.} \cite{fang2020multi} treated the voxels with unknown labels as the background, and then proposed the target adaptive loss (TAL) for a segmentation network that is trained on multiple partially labeled datasets. 
Shi \textit{et al.} \cite{shi2020marginal} merged unlabeled organs with the background and imposed an exclusive constraint on each voxel (\textit{i.e.} each voxel belongs to either one organ or the background) for learning a segmentation model jointly on a fully labeled dataset and several partially labeled datasets. 
To learn multi-class segmentation from single-class datasets, Dmitriev \textit{et al.} \cite{dmitriev2019learning} utilized the segmentation task as a prior and incorporated it into the intermediate activation signal.

The proposed DoDNet is different from these methods in three main aspects:
(1) \cite{fang2020multi,shi2020marginal} formulate the partially labeled issue as a multi-class segmentation task and treat unlabeled organs as the background, which may be misleading since the organ unlabeled in this dataset is indeed the foreground on another task. To address this issue, we formulate the partially labeled problem as a single-class segmentation task, aiming to segmenting each organ respectively;
(2) Most of these methods adopt the multi-head architecture, which is composed of a shared backbone network and multiple segmentation heads for different tasks. Each head is either a decoder \cite{chen2019med3d} or the last segmentation layer \cite{fang2020multi,shi2020marginal}. In contrast, the proposed DoDNet is a single-head network, in which the head is flexible and dynamic;
(3) Our DoDNet uses the dynamic segmentation head to address the partially labeled issue, instead of embedding the task prior into the encoder and decoder; 
(4) Most existing methods focus on multi-organ segmentation, while our DoDNet segments both organs and tumors, which is more challenging. 

{\noindent\bf Dynamic filter learning}
Dynamic filter learning has drawn considerable research attention in the computer vision community due to its adaptive nature \cite{jia2016dynamic,yang2019condconv,chen2020dynamic,Tian2020CondInst,he2019dynamic,HDFNet_ECCV2020}.
Jia \textit{et al.} \cite{jia2016dynamic} designed a dynamic filter network, in which the filters are generated dynamically conditioned on the input. This design is more flexible than traditional convolutional networks, where the learned filters are fixed during the inference.
Yang \textit{et al.} \cite{yang2019condconv} introduced the conditionally parameterized convolutions, which learn specialized convolutional kernels for each input, and effectively increase the size and capacity of a convolutional neural network.
Chen \textit{et al.} \cite{chen2020dynamic} presented another dynamic network, which dynamically generates attention weights for multiple parallel convolution kernels and assembles these kernels to strengthen the representation capability. 
Pang \textit{et al.} \cite{HDFNet_ECCV2020} integrated the features of RGB images and depth images to generate dynamic filters for better use of cross-modal fusion information in RGB-D salient object detection. 
Tian \textit{et al.} \cite{Tian2020CondInst} applied the dynamic convolution to instance segmentation, where the filters in the mask head are dynamically generated for each target instance. 
These methods successfully employ the dynamic filer learning towards certain ends,
such as 
increasing the network flexibility \cite{jia2016dynamic}, enhancing the representation capacity \cite{yang2019condconv, chen2020dynamic}, integrating cross-modal fusion information \cite{HDFNet_ECCV2020}, or abandoning the use of instance-wise ROIs \cite{Tian2020CondInst}. Comparing to these works,
our %
work here %
differs as follows.
1) we employ the dynamic filter learning to address the partially labeling issue for 3D medical image segmentation; and 2) the dynamic filters generated in our DoDNet are conditioned not only on the input image, but also on the assigned task.

\section{Our Approach}
\label{Sec.Approach}

\subsection{Problem definition}
Let us consider $m$ partially labeled datasets $\{ \mathfrak{D}_{1},$ $ \mathfrak{D}_{2},$ $..., \mathfrak{D}_{m} \}$, which were collected for $m$ organ and tumor segmentation tasks:
\[
\{ {\rm S_1:liver\&tumor;   \, \rm  S_2:kidney\&tumor, ...} \}.
\]
Here, $\mathfrak{D}_{i}=\{ \textbf{X}_{ij}, \textbf{Y}_{ij} \}_{j=1}^{n_i}$ represents the $i$-th partially labeled dataset that contains $n_i$ labeled images. 
The $j$-th image in $\mathfrak{D}_{i}$ is denoted by $\textbf{X}_{ij} \in \mathbb{R}^{D\times W \times H}$, where $W \times H$ is the size of each slice and $D$ is number of slices. The corresponding segmentation ground truth is $\textbf{Y}_{ij}$, where the label of each voxel belongs to $\{\rm 0:background; 1:organ; 2:tumor \}$. 
Straightforwardly, this partially labeled multi-organ and tumor segmentation problem can be solved by training $m$ segmentation networks $\{ f_1, f_2, ..., f_m \}$ on $m$ datasets, respectively, shown as follows
\begin{equation}
\left\{\begin{matrix}
&\min_{\bm{\theta}_1} \sum_{j=1}^{n_1} \mathcal{L} (f_1(\textbf{X}_{1j}; \bm{\theta}_1), \textbf{Y}_{1j}) \\ 
&\vdots \\ 
&\min_{\bm{\theta}_m} \sum_{j=1}^{n_m} \mathcal{L} (f_m(\textbf{X}_{mj}; \bm{\theta}_m), \textbf{Y}_{mj}) & 
\end{matrix}\right.
\end{equation}
where $\mathcal{L}$ represents the loss function of each network, $\{ \theta_1, \theta_2, ..., \theta_m \}$ represent the parameters of these $m$ networks. 
In this work, we attempt to address this problem using only one single network $f$, which can be formally expressed as  
\begin{equation}
\min_{\bm{\theta}} \sum_{i=1}^{m} \sum_{j=1}^{n_i} \mathcal{L} (f(\textbf{X}_{ij}; \bm{\theta}), \textbf{Y}_{ij})
\end{equation}
The DoDNet proposed here for this purpose consists of a shared encoder-decoder, a task encoding module, a dynamic filter generation module, and a dynamic segmentation head (see Fig.~\ref{fig:framework}). We now delve into the details of each part.

\subsection{Encoder-decoder architecture}
\label{Sec.encoder-decoder}

The main component of DoDNet is the shared encoder-decoder that has an U-like architecture \cite{ronneberger2015u}. 
The encoder is composed of repeated applications of 3D residual blocks \cite{he2016deep}, each containing two convolutional layers with $3\times3\times3$ kernels. Each convolutional layer is followed by group normalization \cite{wu2018group} and ReLU activation. 
At each downsampling step, the convolution with a stride of 2 is used to halve the resolution of input feature maps.
The number of filters is set to 32 in the first layer, and is doubled after each downsampling step so as to preseve the time complexity per layer \cite{he2016deep}. We totally perform four downsampling operations in the encoder. 
Given an input image $\textbf{X}_{ij}$, the output feature map is
\begin{equation}
\textbf{F}_{ij} = f_E(\textbf{X}_{ij}; \bm{\theta}_{E})
\end{equation}
where $\bm{\theta}_{E}$ represents all encoder parameters.

Symmetrically, the decoder upsamples the feature map to improve its resolution and halve its channel number step by step. At each step, the upsampled feature map is first summed with the corresponding low-level feature map from the encoder, and then refined by a residual block. After upsampling the feature map four times, we have
\begin{equation}
\textbf{M}_{ij} = f_D(\textbf{F}_{ij}; \bm{\theta}_{D})
\end{equation}
where $\textbf{M}_{ij} \in \mathbb{R}^{C\times D\times W \times H}$ is the pre-segmentation feature map, $\bm{\theta}_{D}$ represents all decoder parameters, and the channel number $C$ is set to 8 (see ablation study in Sec.~\ref{Sec.depth8}). 

The encoder-decoder aims to generate $\textbf{M}_{ij}$, which is supposed to be rich-semantic and not subject to a specific task, \textit{i.e.}, containing the semantic information of multiple organs and tumors.

\subsection{Task encoding}
Each partially labeled dataset contains the annotations of only a specific organ and related tumors. This information is a critical prior that tells the model with which task it is dealing and on which region it should focus.
For instance, given an input sampled from the liver and tumor segmentation dataset, the proposed DoDNet is expected to be specialized for this task, \textit{i.e.}, predicting the masks of liver and liver tumors while ignoring other organs and other tumors.
Intuitively, this task prior should be encoded into the model for task-awareness. 
Chen et al. \cite{chen2017fast} encoded the task as a $m$-dimensional one-hot vector, and concatenated the extended task encoding vector with the input image to form an augmented input. Owing to task encoding, the network `awares' the task through the additional input channels and thus is able to accomplish multiple tasks, albeit with the performance degradation. However, the channel of input increases from $1$ to $m+1$, leading to a dramatic increase of computation and memory cost.
In this work, we also encode the task prior of each input $\textbf{X}_{ij}$ into a $m$-dimensional one-hot vector $\textbf{T}_{ij} \in \{0, 1\}^{m}$, shown as follows 

\begin{equation}
{ \textbf{T}}_{ijk} =\left\{
\begin{array}{l l}
0            & {\rm if} \ k \neq i \\
1            & {\rm otherwise}
\end{array}
\right.k=1,2,..., m\end{equation}
Here, $\textbf{T}_{ijk}=1$ means that the annotation of $k$-th task is available for the current input $\textbf{X}_{ij}$.
Instead of extending the size of task encoding vector $\textbf{T}_{ij}$ from $\mathbb{R}^{m\times 1\times 1 \times 1}$ to $\mathbb{R}^{m\times D\times W \times H}$ and using it as $m$ additional input channels \cite{chen2017fast}, we first concatenate $\textbf{T}_{ij}$ with the aggregated features and then use the concatenation for dynamic filter generation.
Therefore, both computational and spatial complexity of our task encoding strategy is significantly lower than that in \cite{chen2017fast} (see Figure~\ref{fig:inference_time}).

\subsection{Dynamic filter generation}

For a traditional convolutional layer, the learned kernels are fixed after training and shared by all test cases. Hence, the network optimized on one task must be less-optimal to others, and it is hard to use a single network to perform multiple organ and tumor segmentation tasks. To overcome this difficulty, we introduce a dynamic filter method to generate the kernels, which are specialized to segment a particular organ and tumors. 
Specifically, a single convolutional layer is used as a task-specific controller $\varphi (\cdot)$. The image feature $\textbf{F}_{ij}$ is aggregated via global average pooling (GAP) and concatenated with the task encoding vector $\textbf{T}_{ij}$ as the input of $\varphi(\cdot)$. Then, the kernel parameters $\bm{\omega}_{ij}$ are generated dynamically conditioned not only on the assigned task $S_{i}$ but also on the input image ${\bf X}_{ij}$ itself, expressed as follows
\begin{equation}
\bm{\omega}_{ij} = \varphi({\rm GAP}({ \textbf{F} }_{ij}) 
||
{ \textbf{T} }_{ij}; \bm{\theta}_{ \varphi})
\label{eq:filters}
\end{equation}
where $\bm{\theta}_{\varphi}$ represents the controller parameters, and 
$||$
represents the concatenation operation. 

\begin{table}
\begin{center}
\small 
\begin{tabular}{c|c|c}
\hline
Conv layer & \#Weights               & \#Bias \\ \hline
1          & $8\times8$ & $8$      \\ \hline
2          & $8\times8$ & $8$      \\ \hline
3          & $8\times2$ & $2$      \\ \hline
Totoal     & \multicolumn{2}{c}{$162$}         \\ \hline
\end{tabular}
\end{center}
\caption{Number of parameters generated by controller $\varphi(\cdot )$.}
\label{tab:para162}
\end{table}

\subsection{Dynamic head}
\label{Sec.DynamaicHead}
During the supervised training, it is worthless to predict the organs and tumors whose annotations are not available. 
Therefore, a light-weight dynamic head is designed to enable specific kernels to be assigned to each task for the segmentation of a specific organ and tumors. The dynamic head contains three stacked convolutional layers with $1\times1\times1$ kernels. 
The kernel parameters in three layers, denoted by $\bm{\omega}_{ij} = \{\bm{\omega}_{ij1}, \bm{\omega}_{ij2}, \bm{\omega}_{ij3} \}$, are dynamically generated by the controller $\varphi(\cdot )$ according to the input image and assigned task (see Eq.~\ref{eq:filters}).

The first two layers have 8 channels, and the last layer has 2 channels, \textit{i.e.}, one channel for organ segmentation and the other for tumor segmentation. 
Therefore, a total of 162 parameters (see Table~\ref{tab:para162} for details) are generated by the controller.
The partial predictions of $j$-th image with regard to $i$-th task is computed as
\begin{equation}
\textbf{P}_{ij} = ((\textbf{M}_{ij} \ast \bm{\omega}_{ij1}) \ast \bm{\omega}_{ij2}) \ast \bm{\omega}_{ij3}
\end{equation}
where $\ast$ represents the convolution, and $\textbf{P}_{ij} $ $\in$ $ \mathbb{R}^{2\times D\times W \times H}$ represents the predictions of organs and tumors.
Although each image requires a group of specific kernels for each task, the computation and memory cost of our light-weight dynamic head is negligible compared to the encoder-decoder (see Sec.~\ref{Sec.SpeedvsAccuracy}).

\subsection{Training and Testing}

For simplicity, we treat the segmentation of an organ and related tumors as two binary segmentation tasks, and jointly use the Dice loss and binary cross-entropy loss as the objective for each task. The loss function is formulated as 
\begin{equation}
\begin{aligned}
\small
\mathcal{L} = &1 - \frac{2  \sum_{i=1}^{V} p_i y_i}{ \sum_{i=1}^{V} (p_i + y_i + \epsilon)} \\
&- \sum_{i=1}^{V}{(y_i \log p_i + (1-y_i)\log(1-p_i))}
\end{aligned}
\end{equation}
where $p_i$ and $y_i$ represent the prediction and ground truth of $i$-th voxel, $V$ is the number of all voxels, and $\epsilon$ is added as a smoothing factor. 
We employ a simple strategy to train DoDNet on multiple partially labeled datasets, \textit{i.e.}, ignoring the predictions corresponding to unlabeled targets.
Taking colon tumor segmentation for example, the result of organ prediction is ignored during the loss computation and error back-propagation, since the annotations of organs are unavailable. 

During inference, the proposed DoDNet is flexible to $m$ segmentation tasks. 
Given a test image, the pre-segmentation feature $\textbf{M}_{ij}$ is extracted from the encoder-decoder network. Assigned with a task, the controller generates the kernels conditioned on the input image and task. The dynamic head powered with the generated kernels is able to automatically segment the organ and tumors as specified by the task. 
In addition, if $m$ tasks are all required, our DoDNet is able to generate $m$ groups of kernels for the dynamic head and to efficiently segment all of $m$ organs and tumors in turn. Compared to the encoder-decoder, the dynamic head is so light that the inference cost of $m$ dynamic heads is almost negligible.

\begin{table}
\small 
\caption{Details about MOTS dataset, including partial labels, available annotations, and number of training and test images. \checkmark means the annotations are available and $\times$ is the opposite.}
\begin{center}
\begin{tabular}{l|c|c|c|c}
\hline
\multirow{2}{*}{Partial-label task} & \multicolumn{2}{c|}{Annotations} & \multicolumn{2}{c}{\# Images} \\ \cline{2-5} 
 & Organ & Tumor & Training & Test \\ \hline
\#1 Liver & \checkmark & \checkmark & 104 & 27 \\ \hline
\#2 Kidney & \checkmark & \checkmark & 168 & 42 \\ \hline
\#3 Hepatic Vessel & \checkmark & \checkmark & 242 & 61 \\ \hline
\#4 Pancreas & \checkmark & \checkmark & 224 & 57 \\ \hline
\#5 Colon & $\times$ & \checkmark & 100 & 26 \\ \hline
\#6 Lung & $\times$ & \checkmark & 50 & 13 \\ \hline
\#7 Spleen & \checkmark & $\times$ & 32 & 9 \\ \hline
Total & - & - & 920 & 235 \\ \hline
\end{tabular}
\end{center}
\label{tab:MOTS_details}
\end{table}

\begin{table*}[h]
\centering
\footnotesize
\begin{minipage}{0.3\linewidth}
\setcaptionwidth{1.8in} 
\centering
\caption{Comparison of dynamic head with different depth (\#layers), varying from 2 to 4.}
\label{tab:ablation_depth}
\begin{tabular}{c|c|c}
\hline
Depth & Average\  Dice & Average\  HD \\ \hline
2     &71.30& \textbf{25.72} \\ \hline
3     &\textbf{71.67} &25.86\\ \hline
4     & 71.63 & 26.07 \\ \hline
\end{tabular} 
\end{minipage}
\begin{minipage}{0.30\linewidth}  
\footnotesize
\setcaptionwidth{1.8in} 
\centering
\caption{Comparison of dynamic head with different width (\#channels), varying from 4 to 8.}
\label{tab:ablation_width}
\begin{tabular}{c|c|c}
\hline
Width & Average Dice & Average HD \\ \hline
4     & 69.79 & 30.40 \\ \hline
8     & \textbf{71.67} & \textbf{25.86}\\ \hline
16    & 71.45 & 26.31\\ \hline
\end{tabular}
\end{minipage}
\begin{minipage}{0.39\linewidth}  
\centering
\caption{Comparison of the effectiveness of different conditions (image feature, task encoding) during the dynamic filter generation.}
\label{tab:ablation_condition}
\footnotesize 
\begin{tabular}{c|c|c|c}
\hline
Image feat. & Task enc. & Average Dice & Average HD \\ \hline
\checkmark  & \checkmark & \textbf{71.67} & \textbf{25.86} \\ \hline
$\times$    & \checkmark & 71.26 & 29.38 \\ \hline
\checkmark  & $\times$   & 51.80 & 79.94 \\ \hline
\end{tabular}
\end{minipage}
\end{table*}

\section{Experiment}
\subsection{Experiment setup}
\noindent\textbf{Dataset}: 
We built a large-scale partially labeled \textbf{M}ulti-\textbf{O}rgan and \textbf{T}umor \textbf{S}egmentation (MOTS) dataset using multiple medical image segmentation benchmarks, including LiTS\cite{bilic2019liver}, KiTS \cite{heller2019kits19}, and Medical Segmentation Decathlon (MSD) \cite{simpson2019large}. 
MOTS is composed of seven partially labeled sub-datasets, involving seven organ and tumor segmentation tasks. 
There are 1155 3D abdominal CT scans collected from various clinical sites around the world, including 920 scans for training and 235 for test. 
More details are given in Table~\ref{tab:MOTS_details}. 
Each scan is re-sliced to the same voxel size of $1.5\times0.8\times0.8mm^3$.

The MICCAI 2015 Multi Atlas Labeling \textbf{B}eyond the \textbf{C}ranial \textbf{V}ault (BCV) dataset \cite{BCV_benchmark} was also used for this study. It is composed of 50 abdominal CT scans,including 30 scans for training and 20 for test. Each training scan is paired with voxel-wise annotations of 13 organs, including the liver, spleen, pancreas, right kidney, left kidney, gallbladder, esophagus, stomach, aorta, inferior vena cava, portal vein and splenic vein, right adrenal gland, and left adrenal gland. This dataset provides a downstream task, on which the segmentation network pre-trained on MOTS was evaluated.

\noindent\textbf{Evaluation metric}: 
The Dice similarity coefficient (Dice) and Hausdorff distance (HD) were used as performance metrics for this study. Dice measures the overlapping between a segmentation prediction and ground truth, and HD evaluates the quality of segmentation boundaries by computing the maximum distance between the predicted boundaries and ground truth.

\noindent\textbf{Implementation details}: 
All experiments were performed on a workstation with two NVIDIA 2080Ti GPUs. 
To filter irrelevant regions and simplify subsequent processing, we truncated the HU values in each scan to the range $[-325, +325]$ and linearly normalized them to $[-1, +1]$.
Owing to the benefits of group normalization \cite{wu2018group}, our model adopts the micro-batch training strategy with a small batch size of 2. To accelerate the training process, we also employed the weight standarization \cite{qiao2019weight} that smooths the loss landscape by standardizing the convolutional kernels. 
The stochastic gradient descent (SGD) algorithm with a momentum of 0.99 was adopted as the optimizer. The learning rate was initialized to 0.01 and decayed according to a polynomial policy ${\rm lr}={\rm lr}_{init}\times(1-k/K)^{0.9}$, where the maximum epoch $K$ was set to 1,000. 
In the training stage, we randomly extracted sub-volumes with the size of $64\times192\times192$ from CT scans as the input. 
In the test stage, we employed the sliding window based strategy and let the window size equal to the size of training patches. 
To ensure a fair comparison, the same training strategies, including the weight standarization, learning rate, optimizer, and other settings, were applied to all competing models. 

\subsection{Ablation study}
We split the 20\% of training scans as validation data to perform the ablation study, which investigates the effectiveness of the detailed design of the dynamic head and dynamic filter generation module. We average the Dice score and HD of 11 organs and tumors (listed in Table~\ref{tab:MOTS_details}) as two evaluation indicators for a fair comparison.

\noindent \textbf{Depth of dynamic head:}
\label{Sec.depth8}
In Table~\ref{tab:ablation_depth}, we compared the performance of the dynamic head with different depths, varying from 2 to 4. The width is fixed to 8, except for the last layer, which has 2 channels. 
It shows that, considering the trade-off between Dice and HD, DoDNet achieves the best performance on the validation set when the depth of dynamic head is set to 3. But the performance fluctuation is very small when the depth increases from 2 to 4. The results indicate the robustness of dynamic head to the varying depth. We empirically set the depth to 3 for this study.

\noindent \textbf{Width of dynamic head:}
In Table~\ref{tab:ablation_width}, we compared the performance of the dynamic head with different widths, varying from 4 to 16. The depth is fixed to 3.
It shows that the performance improves substantially when increasing the width from 4 to 8, but drops slightly when further increasing the width from 8 to 16. It suggest that the performance of DoDNet tends to become stable when the width of dynamic head falls within reasonable range ($\geq 8$). Considering the complexity issue, we empirically set the width of dynamic head to 8.

\noindent \textbf{Condition analysis:}
The kernels of dynamic head are generated on condition of both the input image and assigned task. We compared the effectiveness of both conditions in Table~\ref{tab:ablation_condition}. 
It reveals that the task encoding plays a much more important role than image features in dynamic filer generation. It may be attributed to the fact that the task prior is able to make DoDNet aware of what task is being handled. %
Without the task condition, all kinds of organs, like liver, kidneys, and pancreas, are equally treated as the same foreground. 
In this case, it is hard for DoDNet to fit such a complex foreground that is composed of multiple organs. 
Moreover, DoDNet fails to distinguish each specific organ or tumors from this foreground without the task condition.

\begin{table*}
\small
\caption{Performance (Dice, \%, higher is better; HD, lower is better) of different methods on seven partially labeled datasets. Note that `Average score' is the aggregative indicator that averages the Dice or HD over 11 categories.}
\begin{center}
\begin{tabular}
{c|m{0.76cm}<{\centering}|m{0.76cm}<{\centering}|m{0.76cm}<{\centering}|m{0.76cm}<{\centering}|m{0.76cm}<{\centering}|m{0.76cm}<{\centering}|m{0.76cm}<{\centering}|m{0.76cm}<{\centering}|m{0.76cm}<{\centering}|m{0.76cm}<{\centering}|m{0.76cm}<{\centering}|m{0.76cm}<{\centering}}
\hline
\multirow{3}{*}{Methods} & \multicolumn{4}{c|}{Task 1: Liver} &  \multicolumn{4}{c|}{Task 2: Kidney} & \multicolumn{4}{c}{Task 3: Hepatic Vessel} \\ \cline{2-13} & \multicolumn{2}{c|}{Dice} & \multicolumn{2}{c|}{HD} & \multicolumn{2}{c|}{Dice} & \multicolumn{2}{c|}{HD} & \multicolumn{2}{c|}{Dice} & \multicolumn{2}{c}{HD} \\ \cline{2-13} 
 & Organ & Tumor & Organ & Tumor & Organ & Tumor & Organ & Tumor & Organ & Tumor & Organ & Tumor \\ \hline
Multi-Nets & 96.61 & 61.65 & 4.25 & 41.16 & 96.52 & 74.89 & \color{blue}{1.79} & 11.19 & 
\color{red}{63.04} & 72.19 & 13.73 & 50.70 \\ \hline
TAL \cite{fang2020multi} & 96.18 & 60.82 & 5.99 & 38.87 & 95.95 & 75.87 & 1.98 & 15.36 & 61.90 & 72.68 & 13.86 & \color{blue}{43.57} \\ \hline
Multi-Head \cite{chen2019med3d} & \color{blue}{96.75} & 64.08 & \color{blue}{3.67} & 45.68 & \color{blue}{96.60} & \color{blue}{79.16} & 4.69 & 13.28 & 59.49 & 69.64 & 19.28 & 79.66 \\ \hline
Cond-NO & 69.38 & 47.38 & 37.79 & 109.65 & 93.32 & 70.40 & 8.68 & 24.37 & 42.27 & 69.86 & 93.35 & 70.34 \\ \hline
Cond-Input \cite{chen2017fast} & 96.68 & \color{blue}{65.26} & 6.21 & 47.61 & \color{red}{96.82} & 78.41 & \color{red}{1.32} & 10.10 & 62.17 & \color{blue}{73.17} & \color{blue}{13.61} & \color{red}{43.32} \\ \hline
Cond-Dec \cite{dmitriev2019learning} & 95.27 & 63.86 & 5.49 & \color{red}{36.04} & 95.07 & \color{red}{79.27} & 7.21 & \color{red}{8.02} & 61.29 & 72.46 & 14.05 & 65.57 \\ \hline
DoDNet & \color{red}{96.87} & \color{red}{65.47} & \color{red}{3.35} & \color{blue}{36.75} & 96.52 & 77.59 & 2.11 & \color{blue}{8.91} & \color{blue}{62.42} & \color{red}{73.39} & \color{red}{13.49} & 53.56 \\ \hline \hline
\multirow{3}{*}{Methods} & \multicolumn{4}{c|}{Task 4: Pancreas}               & \multicolumn{2}{c|}{Task 5: Colon} & \multicolumn{2}{c|}{Task 6: Lung} & \multicolumn{2}{c|}{Task 7: Spleen} & \multicolumn{2}{c}{Average score}              \\ \cline{2-13} 
 & \multicolumn{2}{c|}{Dice} & \multicolumn{2}{c|}{HD} & Dice             & HD              & Dice            & HD              & Dice             & HD               & \multirow{2}{*}{Dice$\uparrow$} & \multirow{2}{*}{HD$\downarrow$} \\ \cline{2-11}
 & Organ & Tumor & Organ & Tumor & Tumor & Tumor & Tumor & Tumor & Organ & Organ & & \\ \hline
Multi-Nets & 82.53 & 58.36 & 9.23 & 26.13 & 34.33 & 103.91 & 54.51 & 53.68 & 93.76 & \color{red}{2.65} & 71.67 & 28.95 \\ \hline
TAL \cite{fang2020multi} & 81.35 & 59.15 & 9.02 & 21.07 & 48.08 & 66.42 & 61.85 & 39.92 & 93.01 & \color{blue}{3.10} & 73.35 & \color{blue}{23.56} \\ \hline
Multi-Head \cite{chen2019med3d} & \color{red}{83.49} & \color{red}{61.22} & \color{red}{6.40} & \color{blue}{18.66} & 50.89 & 59.00 & \color{blue}{64.75} & 34.22 & \color{red}{94.01} & 3.86 & 74.55 & 26.22 \\ \hline
Cond-NO & 65.31 & 46.24 & 36.06 & 76.26 & 42.55 & 76.14 & 57.67 & 102.92 & 59.68 & 38.11 & 60.37 & 61.24 \\ \hline
Cond-Input \cite{chen2017fast} & 82.53 & \color{blue}{61.20} & 8.09 & 31.53 & 51.43 & \color{red}{44.18} & 60.29 & 58.02 & 93.51 & 4.32 & \color{blue}{74.68} & 24.39 \\ \hline
Cond-Dec \cite{dmitriev2019learning} & 77.24 & 55.69 & 17.60 & 48.47 & \color{red}{51.80} & 63.67 & 57.68 & 53.27 & 90.14 & 6.52 & 72.71 & 29.63 \\ \hline
DoDNet & \color{blue}{82.64} & 60.45 & \color{blue}{7.88} & \color{red}{15.51} & \color{blue}{51.55} & \color{blue}{58.89} & \color{red}{71.25} & \color{red}{10.37} & \color{blue}{93.91} & 3.67 & \color{red}{75.64} & \color{red}{19.50} \\ \hline
\end{tabular}
\end{center}
\label{tab:SOTA}
\end{table*}

\subsection{Comparing to state-of-the-art methods}

We compared the proposed DoDNet to state-of-the-art methods, which also attempt to address the partially labeling issue, on seven partially labeled tasks using the MOTS test set. 
The competitors include (1) seven individual networks, each being trained on a partially dataset (denoted by Multi-Nets), (2) two multi-head networks (\textit{i.e.}, Multi-Head~\cite{chen2019med3d} and TAL~\cite{fang2020multi}), (3) a single-network method without the task condition (Cond-NO), and (4) two single-network methods with the task condition (\textit{i.e.}, Cond-Input~\cite{chen2017fast} and Cond-Dec~\cite{dmitriev2019learning}).
To ensure a fair comparison, we used the same encoder-decoder architecture for all methods, except that the channels of decoder layers in Multi-Head were halved due to GPU memory limitation. 

Table~\ref{tab:SOTA} shows the performance metrics for the segmentation of each organ $/$ tumor and the average scores over 11 categories. It reveals that
(1) most of methods (TAL, Multi-Head, Cond-Input, Cond-dec, DoDNet) achieve better performance than seven individual networks (Multi-Nets), suggesting that training with more data (even partially labelled) is beneficial to model performance;
(2) Cond-NO fails to segment multiple organs and tumors when the task condition is unavailable, demonstrating the importance of task condition for a single network to address the partially labeling issue (consistent with the observation in Table~\ref{tab:ablation_condition});
(3) the dynamic filter generation strategy is superior to directly embedding the task condition into the input or decoder (used in Cond-Input and Cond-Dec); and
(4) the proposed DoDNet achieves the highest overall performance with an averaged Dice of 75.64\% and an averaged HD of 19.50. 

\begin{figure}[t]
\begin{center}
\includegraphics[width=1.0\linewidth]{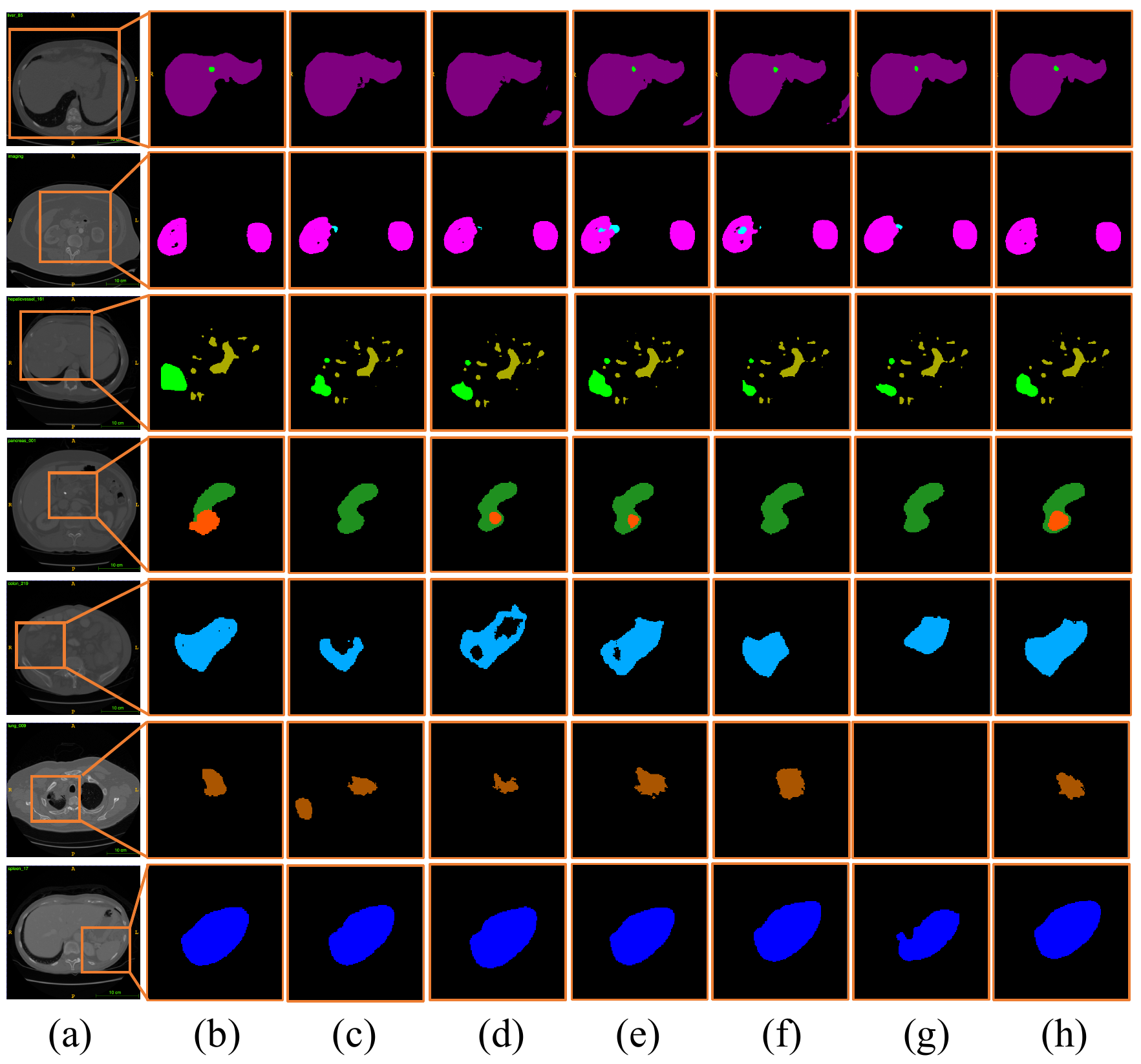}
\end{center}
\caption{Visualization of segmentation results obtained by different methods. (a) input image; (b) ground truth; (c) Multi-Nets; (d) TAL~\cite{fang2020multi}; (e) Multi-Head~\cite{chen2019med3d}; (f) Cond-Input~\cite{chen2017fast}; (g) Cond-Dec~\cite{dmitriev2019learning}; (h) DoDNet. 
}
\label{fig:visual}
\end{figure}

To make a qualitative comparison, we visualized the segmentation results obtained by six methods on seven tasks in Figure~\ref{fig:visual}. It shows that our DoDNet outperforms other methods, especially in segmenting small tumors. 

\begin{figure}[t]
\begin{center}
\includegraphics[width=1.0\linewidth]{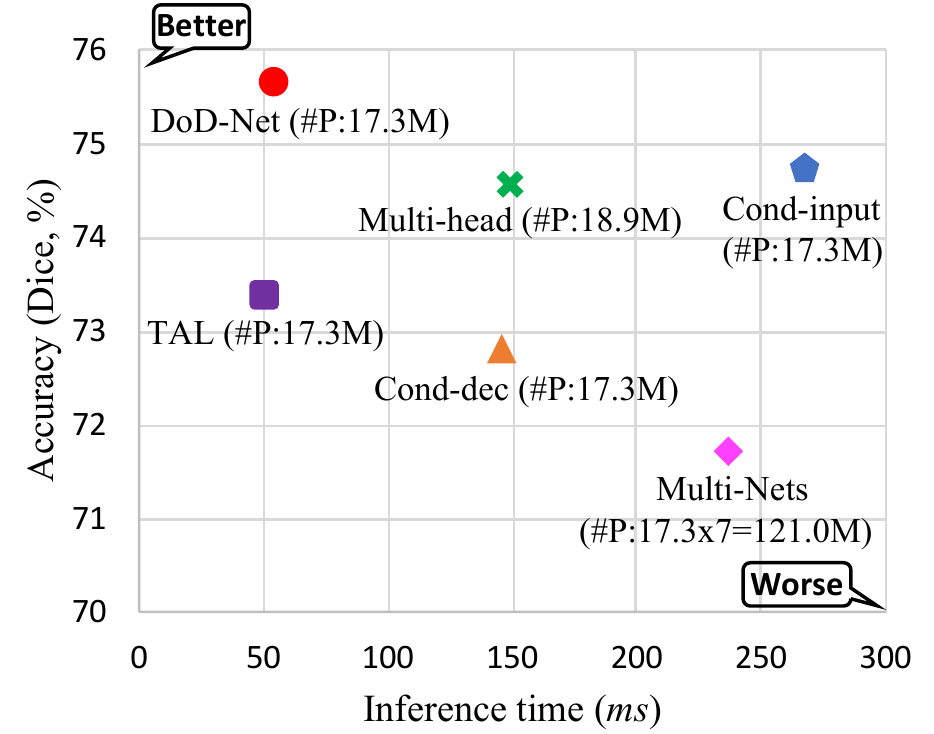}
\end{center}
\caption{Speed vs.\  accuracy. The accuracy refers to the overall Dice score on the MOTS test set. The inference time is computed based on a single input with 64 slices of spatial size $128 \times 128$). `\#P': the number of parameters. `M': Million. 
}
\label{fig:inference_time}
\end{figure}

\label{Sec.SpeedvsAccuracy}
In Figure~\ref{fig:inference_time}, we also compared the speed-accuracy trade-off of five methods. 
Single-network methods, including TAL, Cond-Dec, Cond-Input, and DoDNet, share the encoder and decoder for all tasks, and hence have a similar number of parameters, \textit{i.e.}, 17.3M. Although our DoDNet has an extra controller, the number of parameters in it is negligible.
The Multi-Head network has a little more parameters (\textit{i.e.}, 18.9M) due to the use of multiple task-specific decoders. 
Multi-Nets has to train seven networks to address these partially labeled tasks, resulting in seven times more parameters than a single network.

As for inference speed, Cond-Input, Multi-Nets, Multi-Head, and Cond-Dec suffer from the repeated inference processes, and hence need more time to segment seven kinds of organ and tumors than other methods. 
In contrast, TAL is much more efficient to segment all targets, since the encoder-decoder (excepts for the last segmentation layer) is shared by all tasks.
Our DoDNet shares the encoder-decoder architecture and specializes the dynamic head for each partially labeled task. Due to the light-weight architecture, the inference of dynamic head is very fast. 
In summary, our DoDNet achieves the best accuracy and a fast inference speed.

\subsection{MOTS Pre-training for downstream tasks}
Although achieving startling success driven by large-scale labeled data, deep learning remains trammelled by the limited annotations in medical image analysis. 
The largest partially labeled dataset, \textit{i.e.}, \#3 Hepatic Vessel, only contains 242 training cases, which is much smaller than MOTS with 920 training cases. 
It has been generally recognized that training a deep model with more data contributes to a better generalization ability \cite{zoph2020rethinking,he2019rethinking,tran2018closer,esteva2017dermatologist}.
Therefore, pre-training a model on MOTS should be beneficial for the annotation-limited downstream task.
To demonstrate this, we treated the BCV multi-organ segmentation as a downstream task and conducted experiments on the BCV dataset. We initialized the segmentation network, which has the same encoder-decoder structure as introduced in Sec.~\ref{Sec.encoder-decoder}, using three initialization strategies, including randomly initialization (\textit{i.e.}, training from scratch), pre-training on the \#3 Hepatic Vessel dataset, and pre-training on MOTS.

\begin{figure}[t]
\begin{center}
\includegraphics[width=1.0\linewidth]{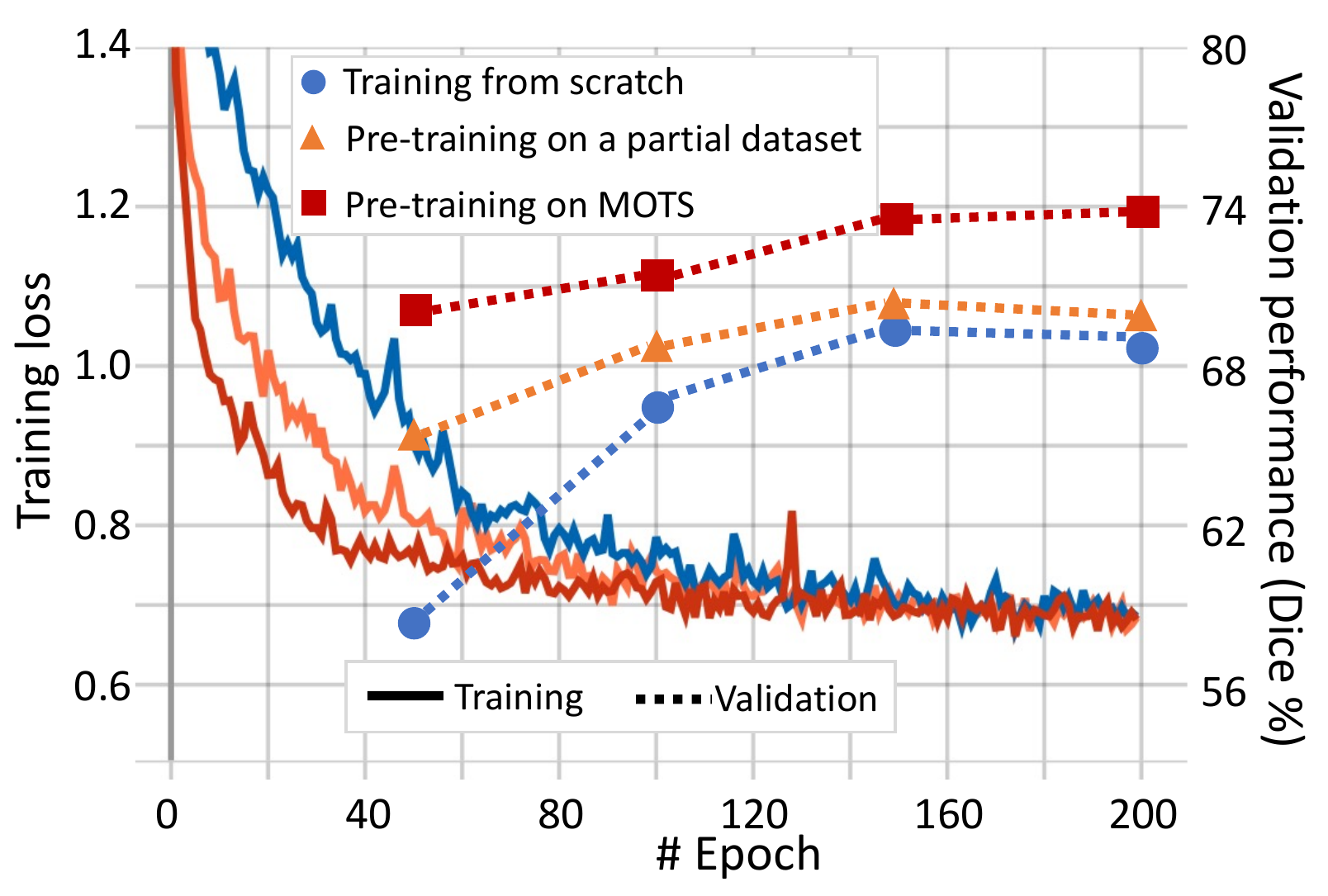}
\end{center}
\caption{Comparison of training loss and validation performance of segmentation network using three initialization strategies, including training from scratch, pre-training on \#3 Hepatic Vessel, and pre-training on MOTS. Here the validation performance refers to the averaged Dice score over 13 categories.
}
\label{fig:BCV_pretrain}
\end{figure}

First, we split 20 cases from the BCV training set for validation, since the annotations of BCV test set were withheld for online assessment, which is inconvenient.
Figure~\ref{fig:BCV_pretrain} shows the training loss and validation performance of the segmentation network with three initialization strategies. The validation performance is measured by the averaged Dice score calculated over 13 categories.
It revels that, comparing to training from scratch, pre-training the network helps it converge quickly and perform better, particularly in the initial stage.
Moreover, pre-training on a small dataset (\textit{i.e.}, \#3 Hepatic Vessel) only slightly outperforms training from scratch, but pre-training on MOTS, which is much larger than \#3 Hepatic Vessel, achieves not only the fastest convergence, but also a remarkable performance boost. The results demonstrate the strong generalization ability of the model pre-trained on MOTS. 

Second, we also evaluated the effectiveness of the MOTS pre-trained weights on the BCV unseen test set. 
We compared our method to other state-of-the-art methods in Table~\ref{tab:BCV_test}, including Auto Context~\cite{roth2018multi}, DLTK~\cite{pawlowski2017dltk}, PaNN~\cite{zhou2019prior}, and nnUnet~\cite{isensee2019automated}. 
Comparing to training from scratch, using the MOTS pre-trained weights contributes to a substantial performance gain, improving the average Dice from 85.30\% to 86.44\%, reducing the average mean surface distance (SD) from 1.46 to 1.17, and reducing the average HD from 19.67 to 15.62.
With the help of MOTS pre-training weights, our method achieves the best SD and HD, and second highest Dice on the test set. %

\begin{table}
\caption{Comparison of state-of-the-art methods on the BCV test set. SD: Mean surface distance (lower is better); TFS: Training network from scratch; MOTS: Pre-training on MOTS. The values of three metrics were averaged over 13 categories.}
\begin{center}
\small 
\begin{tabular}{c|c|c|c}
\hline
\small 
Methods & Avg. Dice & Avg. SD & Avg. HD \\ \hline
Auto Context \cite{roth2018multi} & 78.24 & 1.94 & 26.10 \\ \hline
DLTK \cite{pawlowski2017dltk} & 81.54 & 1.86 & 62.87 \\ \hline
PaNN \cite{zhou2019prior} & 84.97 & 1.45 & 18.47 \\ \hline
nnUnet \cite{isensee2019automated} & \textbf{88.10} & 1.39 & 17.26 \\ \hline
TFS & 85.30 & 1.46 & 19.67 \\ \hline
MOTS  & 86.44 & \textbf{1.17} & \textbf{15.62} \\ \hline
\end{tabular}
\end{center}
\label{tab:BCV_test}
\end{table}

\section{Conclusion}

In this paper, we proposed DoDNet, a single encoder-decoder network with a dynamic head, to address the partially labelling issue for multi-organ and tumor segmentation in abdominal CT scans. 
We created a large-scale partially labeled dataset called MOTS and conducted extensive experiments on it. Our results indicate that, \textit{thanks to task encoding and dynamic filter learning, the proposed DoDNet achieves not only the best overall performance on seven organ and tumor segmentation tasks, but also higher inference speed than other competitors.}
We also demonstrated the value of DoDNet and the MOTS dataset by successfully transferring the weights pre-trained on MOTS to downstream tasks for which only limited annotations are available. It suggests that the byproduct of this work (\textit{i.e.}, a pre-trained 3D network) is conducive to other small-sample 3D medical image segmentation tasks.

{\small
\bibliographystyle{ieee_fullname}
\bibliography{egbib}
}

\end{document}